\documentclass[11pt]{article}

\usepackage[preprint]{acl}

\usepackage{times}
\usepackage{latexsym}

\usepackage[T1]{fontenc}
\usepackage[utf8]{inputenc}

\usepackage{microtype}

\usepackage{inconsolata}

\usepackage{graphicx}

\usepackage{adjustbox}
\usepackage{amsmath} 
\usepackage{bbm}
\usepackage{amsfonts}

\usepackage{graphicx}
\usepackage{multirow}
\usepackage[normalem]{ulem}
\useunder{\uline}{\ul}{}
\usepackage{subcaption}
\usepackage{pdfpages}
\usepackage{url}

\title{Auto-Discovery-Bench: Diagnosing Structured State Tracking in Oracle-Guided Discovery} 

\author{
 \textbf{Tingting Chen\textsuperscript{1}},
 \textbf{Beibei Lin\textsuperscript{1}},
 \textbf{Srinivas Anumasa\textsuperscript{1}},
 \textbf{Vedant Shah\textsuperscript{2}},
 \textbf{Zifeng Yuan\textsuperscript{1}},
\\
 \textbf{Qiran Zou\textsuperscript{1}},
 \textbf{Anirudh Goyal\textsuperscript{3}},
 \textbf{Dianbo Liu\textsuperscript{1}},
\\
\\
 \textsuperscript{1}National University of Singapore
 \textsuperscript{2}Mila-Quebec AI institute
 \textsuperscript{3}Meta Superintelligence Labs
\\
 \href{mailto:tingting.c@u.nus.edu@domain}{tingting.c@u.nus.edu}
}

\begin{document}
\maketitle

\begin{abstract}
Interactive discovery requires agents to maintain and update structured beliefs over many rounds of feedback. 
Before evaluating agents in noisy, open-ended scientific environments, it is useful to isolate this prerequisite capability under controlled conditions.
We introduce Auto-Discovery-Bench, a deterministic oracle-guided diagnostic benchmark in which agents recover hidden structures through repeated hypothesis--intervention--feedback cycles. 
The benchmark instantiates three controlled discovery abstractions: directed graph discovery, undirected relational discovery, and symbolic equation discovery.
Across models, performance degrades as the number of variables, trajectory length, and distractors increase. 
A separate trajectory-tracking diagnostic shows that many failures persist even when intervention selection and hypothesis generation are removed, suggesting that limitations in maintaining and integrating long-range structured information are an important bottleneck for oracle-guided discovery.
Auto-Discovery-Bench is not intended to replace realistic discovery environments; rather, it provides a reproducible, low-confound diagnostic testbed for isolating a prerequisite capability for interactive scientific agents.
\end{abstract}

\section{Introduction}

\begin{figure}[t]
\centering

\begin{subfigure}[t]{0.95\linewidth}
\centering
\includegraphics[width=\linewidth]{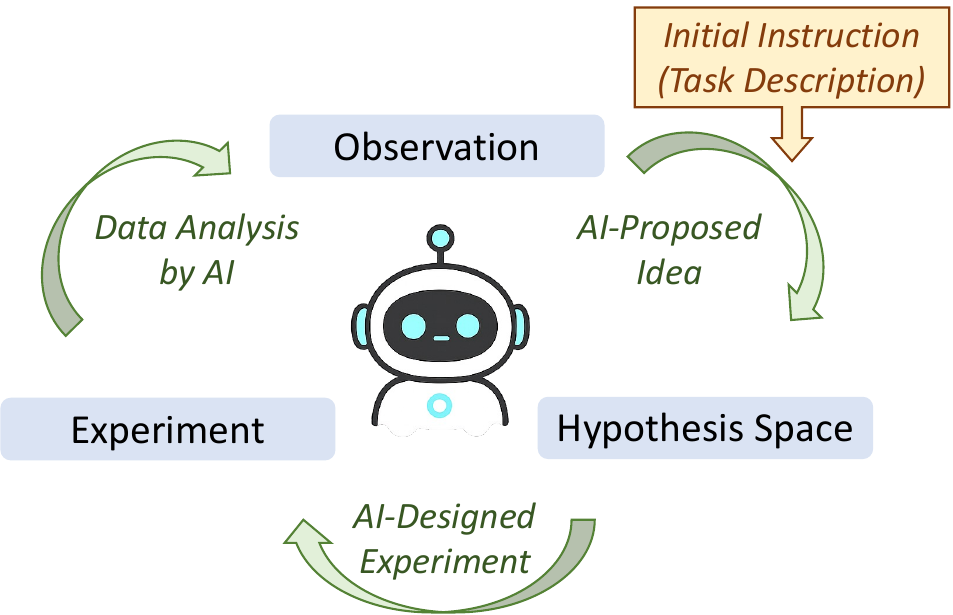}
\caption{Autonomous discovery cycle.}
\label{cycle}
\end{subfigure}

\vspace{1mm}

\begin{subfigure}[t]{0.95\linewidth}
\centering
\includegraphics[width=\linewidth]{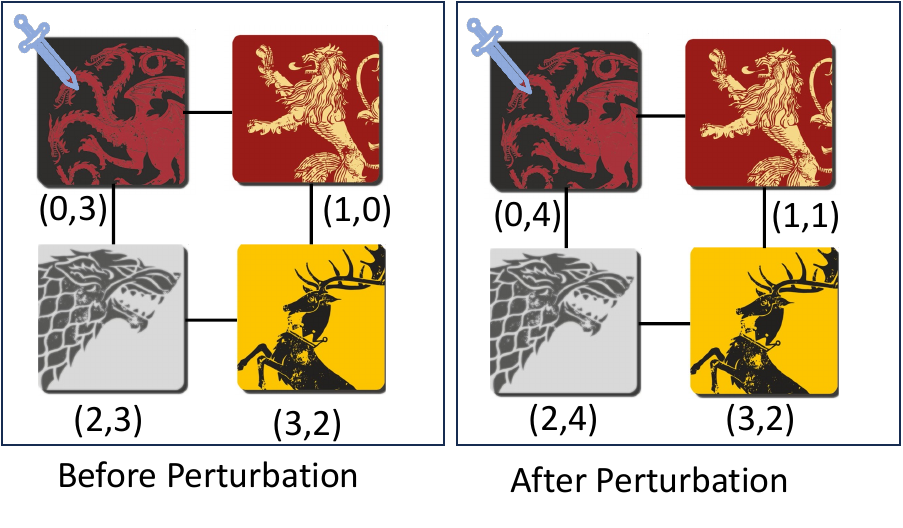}
\vspace{-7mm}
\caption{Example of undirected relational discovery.}
\label{Social}
\end{subfigure}
\vspace{-2mm}
\caption{\textbf{Auto-Discovery-Bench evaluates oracle-guided discovery.}
\textbf{(a) Autonomous Discovery Cycle:} The model iteratively analyzes history, forms a hypothesis $h_t$, proposes an experiment $a_t$, and updates its belief from oracle feedback $o_t$.
\textbf{(b) Undirected relational discovery:} The agent infers hidden edges by perturbing a node (sword icon) and observing how state changes propagate across the graph (e.g., $(2,3)\!\to\!(2,4)$).}
\label{fig:overview}
% \label{cycle}
\end{figure}

Large language models (LLMs) have achieved strong performance across many language benchmarks \cite{xsum,rajpurkar2016squad,wang2018glue}, motivating increasing interest in using them to support scientific workflows, from literature synthesis to hypothesis generation \cite{lu2024ai}.
However, many scientific tasks require iterative, feedback-driven experimentation: observations motivate hypotheses; experiments test them; results update hypotheses and inform subsequent experimental design (Fig.~\ref{cycle}).
This raises a central question: can LLMs operate in an interactive discovery setting—proposing hypotheses, selecting informative experiments, and updating beliefs under feedback—to recover hidden ground truths?
%%%%%%%%% Cycle + Social Figure %%%%%%%%%
%%%%%%%%%%%%%%%%%%%%%%%%%%%%%%%%%%%%
%%%%%%%%% Framework Figure %%%%%%%%%
\begin{figure*}[t]
\centering
{\includegraphics[width=0.95\textwidth]{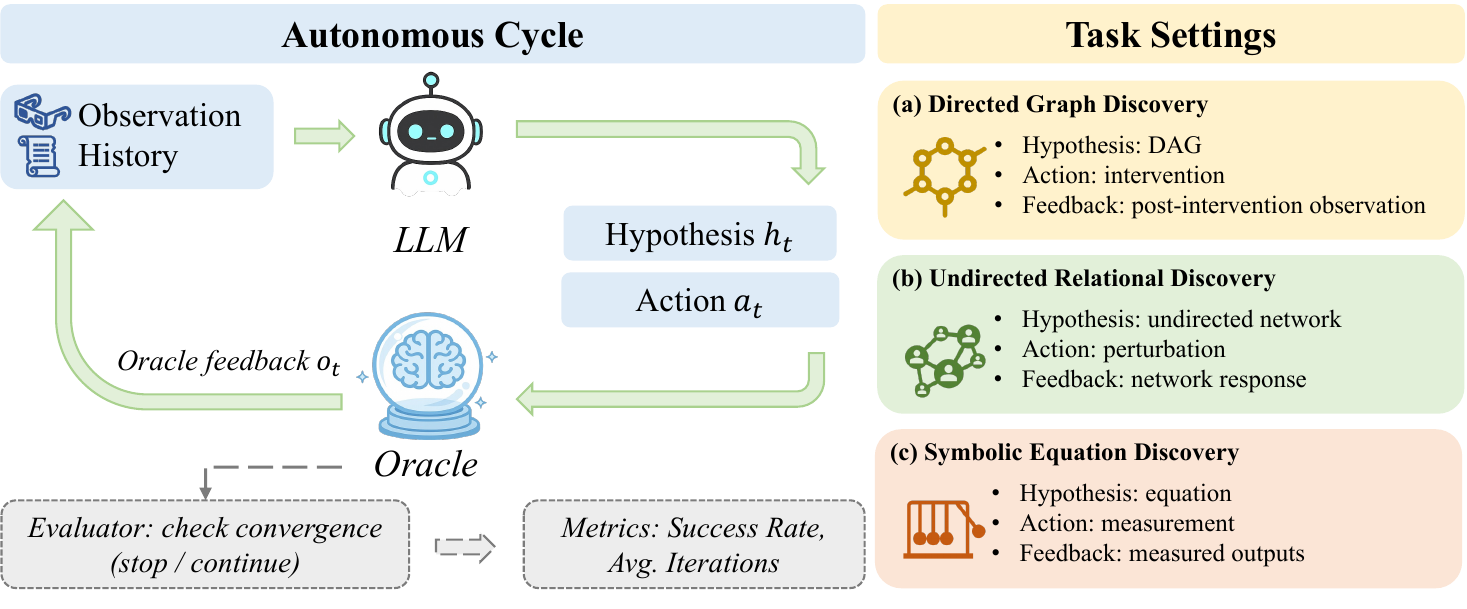}
}\\
\vspace{-2mm}
\caption{\textbf{Overview of Auto-Discovery-Bench.} \textit{Left:} Oracle-guided autonomous cycle. At round $t$, the LLM conditions on the observation history and proposes a hypothesis $h_t$ together with an action $a_t$ (experiment). The oracle executes $a_t$ and returns feedback $o_t$, which is appended to the history for the next round. An evaluator checks convergence and terminates the episode when the stopping criterion is met (or a maximum round budget is reached); we report success rate and average number of rounds. \textit{Right:} Task archetypes. 
(a) Directed graph discovery, where $h_t$ is a DAG and $a_t$ is an intervention; feedback is a post-intervention observation. 
(b) Undirected relational discovery, where $h_t$ is an undirected graph and $a_t$ is a perturbation; feedback is the observed graph response. 
(c) Symbolic equation discovery, where $h_t$ is an equation and $a_t$ specifies a measurement; feedback consists of measured outputs.}
\label{Framework}
\end{figure*}
%%%%%%%%%%%%%%%%%%%%%%%%%%%%%%%%%%%%

Many existing benchmarks evaluate isolated research skills, such as writing and coding \cite{altmae2023artificial,dinu2024symbolicai}, or domain-specific endpoints in material discovery \cite{merchant2023scaling,pyzer2022accelerating} and synthetic biology \cite{jumper2021highly,hayes2025simulating}.
Although recent interactive environments study more realistic discovery behavior, failures in such settings can be difficult to attribute because they conflate exploration, domain knowledge, language grounding, stochastic feedback, and memory.
As a result, there remains a need for controlled diagnostics that preserve the closed-loop structure of discovery while isolating specific prerequisite capabilities.

Our benchmark occupies a middle ground between realistic scientific discovery environments and isolated capability tests. 
Realistic environments such as open-ended laboratory or virtual-world benchmarks offer ecological validity, but failures in such settings can arise from many entangled factors: exploration strategy, domain knowledge, stochastic observations, tool use, language grounding, or memory. At the other extreme, single-shot diagnostic tasks can isolate narrow skills but do not preserve the closed-loop structure of discovery. Our goal is therefore not to simulate real chemistry, sociology, or physics in full fidelity, but to construct controlled discovery abstractions that preserve the iterative hypothesis--experiment--update loop while removing confounds. This allows us to ask a sharper question: when ground truth is deterministic and feedback is clean, can current LLM agents maintain and update structured beliefs over long interaction horizons?

To address this gap, we introduce \textit{Auto-Discovery-Bench}, a controlled oracle-guided diagnostic benchmark for iterative discovery. In each task, the ground truth (e.g., a graph or an equation) is hidden; the model alternates between proposing a hypothesis and selecting an experiment, and an oracle returns feedback after each step.
We instantiate this protocol in three tasks: directed graph discovery, undirected relational discovery, and symbolic equation discovery. Fig.~\ref{Social} provides a concrete example where the agent must infer hidden relationships (edges) between entities (e.g., Game of Thrones houses). The agent performs a perturbation (the sword icon), changing the state of the target node and propagating to its neighbors. By tracking these state shifts (e.g., $(2,3) \to (2,4)$) over multiple rounds, the agent recovers the hidden interaction graph. 

Empirically, we observe large gaps across models as difficulty increases. In the graph-based tasks, many models degrade substantially when scaling from small graphs to 10-node graphs: some require more rounds to converge, while others fail within the interaction budget. To probe the underlying failure mode, we additionally introduce a trajectory-tracking diagnostic, where models must integrate state changes over extended sequences. Accuracy decreases with horizon length, suggesting that limitations in maintaining and integrating long-range structured information are an important bottleneck for oracle-guided discovery.

This paper makes the following contributions:
\begin{itemize}
\item We introduce Auto-Discovery-Bench, a controlled oracle-guided diagnostic benchmark for closed-loop discovery behavior, designed to evaluate structured belief maintenance under repeated feedback while retaining the hypothesis--intervention--update loop.
\item We instantiate the protocol in three task archetypes (directed graph discovery, undirected relational discovery, and symbolic equation discovery) with exact ground truth and deterministic evaluation.
\item We show that model performance and interaction efficiency degrade with larger structures, longer trajectories, and distractor variables, indicating that clean feedback alone is insufficient for reliable discovery.
\item We provide evidence that long-horizon structured state tracking is an important bottleneck, while acknowledging that planning, representation format, and prompt sensitivity may also contribute.
\end{itemize}

\section{Related Work}

Many LLM benchmarks evaluate single-shot reasoning through static queries, including arithmetic and symbolic reasoning tasks \cite{ASDiv,GSM8K,SVAMP,mawps}. 
While useful for measuring local reasoning, these tasks do not preserve the closed-loop structure of discovery, where observations inform later interventions and hypotheses. 
Our work is also related to LLM-based causal graph discovery \cite{jiralerspong2024efficient,long2023can,choi2022lmpriors,kiciman2023causal}, which often queries edge existence or relies on variable metadata. 
In contrast, Auto-Discovery-Bench requires agents to actively choose interventions, receive oracle feedback, and update structured hypotheses over multiple rounds.

Recent benchmarks study richer interactive discovery behavior, including open-ended scientific environments such as DiscoveryWorld \cite{jansen2024discoveryworld} and controlled physics-discovery settings such as PhysGym \cite{chen2025physgym}. 
PhysGym is closest to our symbolic equation discovery setting, but its main focus is controlled prior knowledge in physics environments, obtained by masking context, variable descriptions, and variable names. 
Auto-Discovery-Bench is complementary: it studies long-horizon structured belief maintenance across multiple controlled discovery archetypes and includes a trajectory-tracking diagnostic that isolates state tracking from exploration and hypothesis generation.

%%%%%%%%% Chemistry Figure %%%%%%%%%
\begin{figure}[t]
\centering
{\includegraphics[width=0.95\linewidth]{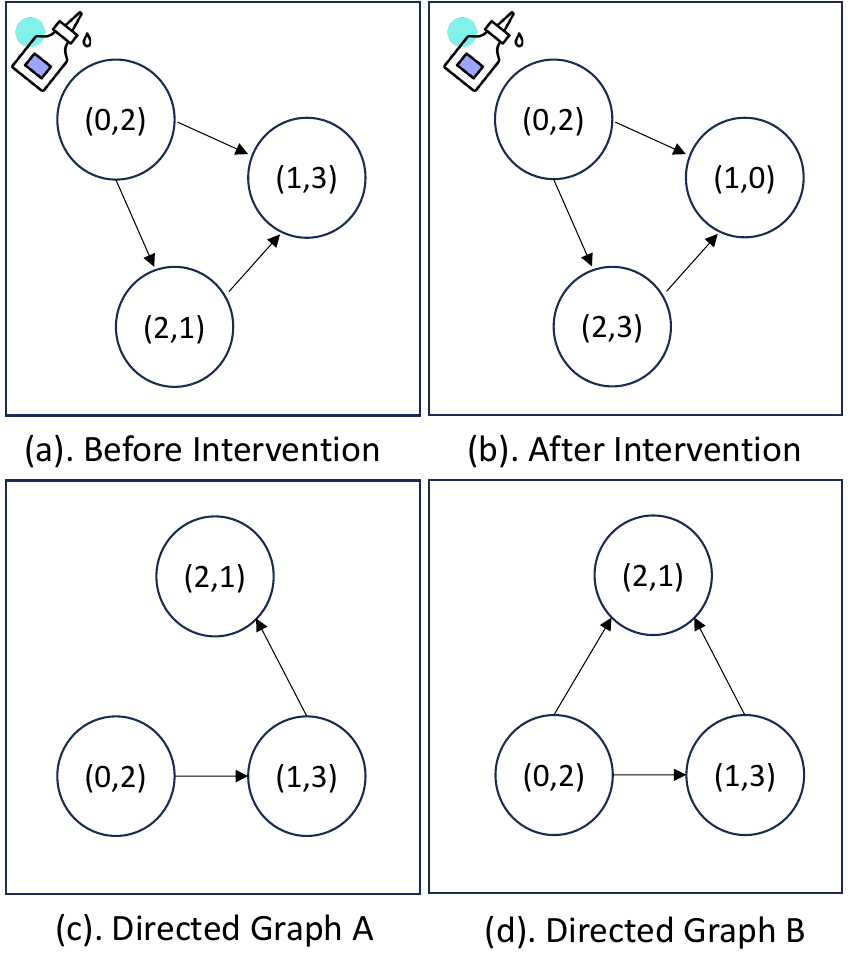}
}\\
\vspace{-2mm}
\caption{Illustration of the directed graph discovery. The brackets indicate (molecule index, molecule state). Figures (a) and (b) illustrate the change in state after an intervention on molecule 0. Figures (c) and (d) present a case where directed graph A and directed graph B result in the same observations.}
\label{Chemistry}
\end{figure}
%%%%%%%%%%%%%%%%%%%%%%%%%%%%%%%%%%%%

\section{Auto-Discovery-Bench}
\label{Methodology}
\subsection{Scope and Design Philosophy}
Auto-Discovery-Bench is designed around three principles.

First, we prioritize diagnostic control over ecological realism. 
All tasks have deterministic oracles, known ground truth, and exact evaluation metrics. 
This makes failures easier to attribute than in open-ended environments where language grounding, domain knowledge, stochasticity, and tool-use errors are entangled.

Second, the benchmark preserves the closed-loop structure of discovery. 
Unlike single-shot matrix or equation tasks, the agent must repeatedly choose an action, observe feedback, and revise a structured hypothesis under an interaction budget.

Third, the benchmark is intended to expose lower-bound failures. 
If a model cannot maintain structured state in clean, fully observed, noise-free settings, then similar failures are likely to become more severe in realistic settings with noise, partial observability, and high-dimensional observations.

Based on these principles, we introduce three diagnostic tasks: (i) directed graph discovery, (ii) undirected relational discovery, and (iii) symbolic equation discovery.

\subsection{Directed Graph Discovery}
\label{Chemistry Environment}
In this setting, we use directed acyclic graphs (DAGs) as a controlled abstraction of directed dependency structures, as illustrated in Fig.~\ref{Chemistry}(a) and (b). 
Each node is described as a molecule for concreteness, and directed edges indicate dependency relations. For each node, a state value is assigned randomly from the set $\{0, ..., S\}$, where $S$ is the maximum state value. An intervention on a molecule causes all children molecules to change their state to a random state, while the state of the intervened molecule remains unaffected. 

Let $H_{ch} \in \{0,1\}^{N \times N}$ be the adjacency matrix representing the relationships between molecules, where  $N$ denotes the number of molecules. Each element of the matrix is either 1 or 0, with 1 indicating the presence of a directed edge and 0 indicating its absence. The ultimate goal for the LLMs is to infer this underlying matrix $H_{ch}$. 

We begin our experiment with an initial prompt that encapsulates all the details such as the problem statement and ultimate goal, and prompts the model to come up with an intervention to learn more about the DAG. More details are provided in the Appendix~\ref{Appendix}. In summary, the model is given the current node states and is prompted to suggest an intervention. After receiving a response from the model, based on the suggested intervention, the oracle executes the intervention and provides the new observations. In the consecutive cycles, the prompt includes the previous interventions performed by the LLM and all observations corresponding to those interventions in the form of an observation matrix. We denote the observation matrix as $G_{ch} \in \{0, ..., S\}^{M \times N}$, where $M$ represents the number of cycles and $N$ represents the number of molecules. The cycles continue until $K_{ch}$ matches the ground-truth $H_{ch}$ or reaches a predefined number of cycles.   

\subsubsection{Evaluation} For a given DAG, multiple adjacency matrices can exhibit similar causal behavior. As shown in Fig.~\ref{Chemistry}(c) and (d), even though the graph structures differ, interventions may yield similar effects on the remaining nodes. Therefore, it is essential to consider this property when evaluating the similarity between generated matrices and the ground truth. 
Let \(K_{ch}^{n}\) denote the matrix \(K_{ch}\) raised to the \(n^{\text{th}}\) power, i.e., \(K_{ch}\) multiplied by itself \(n\) times. 
For all $i,j$, we define
\begin{equation}
H(i,j)=
\begin{cases}
1, & \sum_{n=1}^{M} K_{ch}^{n}(i,j) > 0,\\
0, & \text{otherwise.}
\end{cases}
\end{equation}

Given the hypothesis matrix produced by the model, we compute the corresponding reachability matrix for both the hypothesis and ground truth matrices. Then the resultant matrices are compared to each other to evaluate the similarities. 

\subsection{Undirected Relational Discovery}
\label{Social Networks}
Similar to our directed graph discovery, relationships between individuals are represented as graphs, as illustrated in Fig.~\ref{Social}. The main differences lie in the structure of the connections between nodes and in how they influence each other. To make the task more intuitive, we use the houses in Game of Thrones as an example. 

The connections between houses are undirected. Here, an intervention represents an attack on a specific house, while the house state reflects its anger level. An intervention on a particular house increases its state, as well as the states of its neighboring houses by 1. Consequently, the adjacency matrix $H_{so} \in \{0,1\}^{N \times N}$ should be symmetric, where $N$ represents the number of houses in the relational network. The observation matrix here is denoted as $G_{so} \in \{0, 1, 2...\}^{M \times N}$, where $M$ represents the number of cycles, $N$ represents the number of houses.

%%%%%%%%% Exp Table %%%%%%%%%
\begin{table*}[t!]
\centering
\caption{Performance comparison of multiple LLMs. We present the average number of iterations each LLM requires to obtain the correct answer, along with the success rate.}
\begin{adjustbox}{max width=0.9\textwidth}
\begin{tabular}{ccccccc}
\hline
\multirow{2}{*}{\begin{tabular}[c]{@{}c@{}}Benchmarking\\ Setting\end{tabular}} & \multicolumn{6}{c}{Model used for evaluation (Average Iterations \textbar\ Success Rate)}           \\ \cline{2-7} 
                                                                        & Claude-3-5-haiku       & Llama-3.1-70b-instruct & GPT-4o             & Qwen2.5-72b-instruct & Grok-4 & Gemini-2.5-pro \\ \hline\hline
\multicolumn{7}{c}{\textbf{Task 1: Directed Graph Discovery}} \\ \hline    
\begin{tabular}[c]{@{}c@{}}Nodes: 3; States:3\end{tabular} & \textbf{3.40} \textbar\ 25\% & 5.00 \textbar\ 15\% & 4.15 \textbar\ \textbf{100\%} & 4.15 \textbar\ \textbf{100\%} & 4.22 \textbar\ 90\% & 4.22 \textbar\ 90\% \\
\begin{tabular}[c]{@{}c@{}} Nodes: 3; States: 5\end{tabular}  & 6.00 \textbar\ 5\%          & 4.71 \textbar\ 35\%     & \textbf{4.35 \textbar\ 100\%}& 4.45 \textbar\ \textbf{100\%} & 4.67 \textbar\ 90\% & 4.56 \textbar\ 90\% \\
\begin{tabular}[c]{@{}c@{}} Nodes: 10; States: 5\end{tabular} & $\infty$\textbar\ 0\%   & $\infty$ \textbar\ 0\%        & \textbf{10.67} \textbar\ 15\% & $\infty$ \textbar\ 0\% & 11 \textbar\ \textbf{100\%} & 11.5 \textbar\ \textbf{100\%}    \\ \hline
\multicolumn{7}{c}{\textbf{Task 2: Undirected Relational Discovery}} \\ \hline                              
\begin{tabular}[c]{@{}c@{}}Houses: 3\end{tabular}   & \textbf{1.00} \textbar\ 20\%      & 3.00 \textbar\ 70\%             & 2.90 \textbar\ \textbf{100\%}& 2.00 \textbar\ \textbf{100\%} & 3.10 \textbar\ \textbf{100\%} & 3.40 \textbar\ \textbf{100\%}\\       
\begin{tabular}[c]{@{}c@{}}Houses: 5\end{tabular}   & $\infty$ \textbar\ 0\%   & 6.67 \textbar\ 30\%              & \textbf{4.80 \textbar\ 100\%}& 5.30 \textbar\ \textbf{100\%}  & 5.00 \textbar\ \textbf{100\%} & 5.30 \textbar\ \textbf{100\%} \\
\begin{tabular}[c]{@{}c@{}}Houses: 10\end{tabular}  & $\infty$ \textbar\ 0\%   & $\infty$ \textbar\ 0\%        & \textbf{8.33} \textbar\ 30\% & $\infty$ \textbar\ 0\%  & 10.40 \textbar\ \textbf{100\%} & 10.10 \textbar\ \textbf{100\%}   \\ 
\hline
\multicolumn{7}{c}{\textbf{Task 3: Symbolic Equation Discovery}} \\ \hline 
\begin{tabular}[c]{@{}c@{}}Var: 3; Comp: Simple; Dis:0 \end{tabular} & 4.33 \textbar\ 30\% & 2.00 \textbar\ 10\% & 2.33 \textbar\ 30\% & 2.00 \textbar\ 30\% & \textbf{1.00 \textbar\ 100\%} & \textbf{1.00 \textbar\ 100\%} \\
\begin{tabular}[c]{@{}c@{}}Var: 3; Comp: Medium; Dis:0\end{tabular}  & 2.00 \textbar\ 10\%          & 3.50 \textbar\ 20\%     & 2.67 \textbar\ 30\% & $\infty$ \textbar\ 0\% & \textbf{1.00 \textbar\ 100\%} & \textbf{1.00 \textbar\ 100\%} \\
\begin{tabular}[c]{@{}c@{}} Var: 3; Comp: Complex; Dis:0\end{tabular} & 2.00 \textbar\ 10\%   & 3.00 \textbar\ 10\%        & 2.00 \textbar\ 10\% & 2.00 \textbar\ 10\% & \textbf{1.30 \textbar\ 100\%} & 2.10 \textbar\ \textbf{100\%}    \\
\begin{tabular}[c]{@{}c@{}} Var: 3; Comp: Complex; Dis:2\end{tabular} & 7.00 \textbar\ 10\%   &  $\infty$ \textbar\ 0\% & $\infty$ \textbar\ 0\% & $\infty$ \textbar\ 0\% & \textbf{1.40 \textbar\ 100\%} & 1.89 \textbar\ 90\%    \\ 
\begin{tabular}[c]{@{}c@{}} Var: 5; Comp: Complex; Dis:2\end{tabular} & $\infty$ \textbar\ 0\%  & $\infty$ \textbar\ 0\% & $\infty$ \textbar\ 0\% & $\infty$ \textbar\ 0\% & \textbf{3.22 \textbar\ 90\%} & 4.00 \textbar\ 60\%    \\ \hline
\end{tabular}
\end{adjustbox}
% \vspace{-3mm}
\label{Benchmark}
\end{table*}
%%%%%%%%%%%%%%%%%%%%%%%%%%%%%%%%%%%%

\subsection{Symbolic Equation Discovery}
In this benchmark, we evaluate whether an LLM can recover a hidden symbolic equation through iterative measurements. 
The setting is physics-inspired but intentionally simplified: variables have generic physical names, the ground-truth formula is generated from a controlled grammar, and evaluation is based on exact input-output agreement.

\subsubsection{Task Setup}
In this setting, we simulate physical systems where a target quantity $F$ (force) depends on a subset of available variables: mass ($M$), acceleration ($A$), velocity ($V$), surface area ($S$), temperature ($T$), and energy ($E$). Each variable takes integer values from the range $\{1, ..., 10\}$. The ground-truth equation $H_{ph}$ is hidden from the model. 

At each round $t$, the LLM receives input-output observations $\{(x_i,F_i)\}_{i=1}^{N_{\text{obs}}}$, where $x_i$ represents variable assignments and $F_i$ is the corresponding output. History of previously proposed formulas, accuracy feedback and results from previous actions are also given.
It proposes a hypothesis equation $h_t$ and a measurement action $a_t$ by assigning values to all active variables. 
The oracle scores $h_t$ on random test cases, executes $a_t$, and returns both predicted and true outputs, enabling hypothesis refinement.

\subsubsection{Formula Generation}

To ensure diverse and scalable evaluation, we implement a random formula generator that produces equations of varying complexity: (1) \textbf{Simple:} Pure linear combinations or products without exponents (e.g., $F = M + A + V$, $F = M \cdot A \cdot V$) (2) \textbf{Medium:} Mixed operations with single squared terms (e.g., $F = M^2 \cdot A + V$, $F = M \cdot A - V$) (3) \textbf{Complex:} Multiple squared terms or division operations (e.g., $F = M^2 + A^2 \cdot V$, $F = (M \cdot A) / V$)

Additionally, we introduce distractor variables that appear in the observations but do not affect the true formula. This tests whether models can identify irrelevant variables and avoid overfitting.

\subsubsection{Evaluation}
We evaluate formula correctness by comparing predictions across 100 random test cases with an error tolerance of $\epsilon = 10^{-4}$. A formula is considered correct only if all outputs are identical to the ground truth across all test cases.

\section{Experiments} 
%%%%%%%%% COT Figure %%%%%%%%%%%%
\begin{figure*}[t]
    \centering
    \begin{subfigure}{0.49\textwidth}
        \centering
        \includegraphics[width=\linewidth]{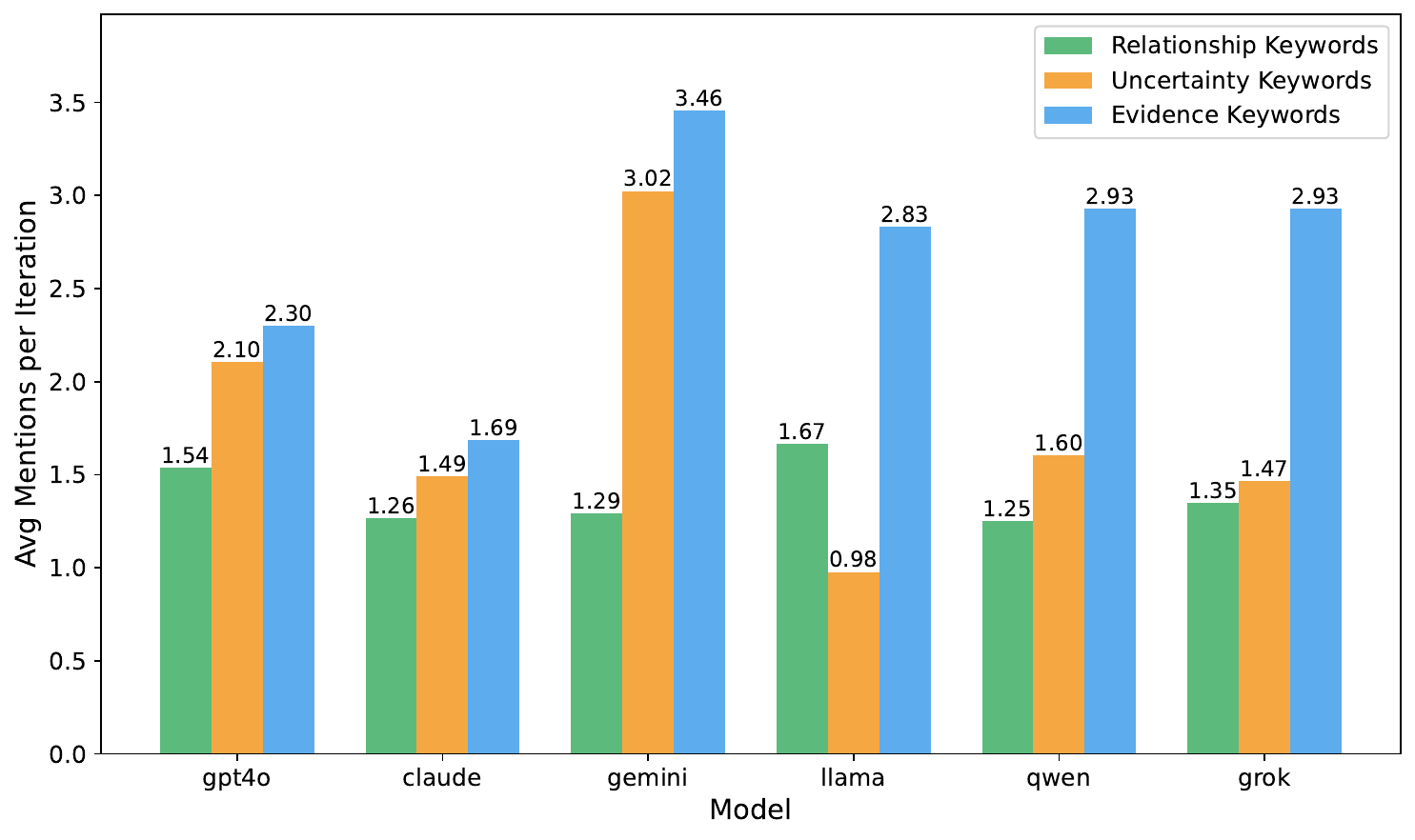}
        \caption{Keyword Mentions per Round}
        \label{keyword}
    \end{subfigure}
    \hfill
    \begin{subfigure}{0.49\textwidth}
        \centering
        \includegraphics[width=\linewidth]{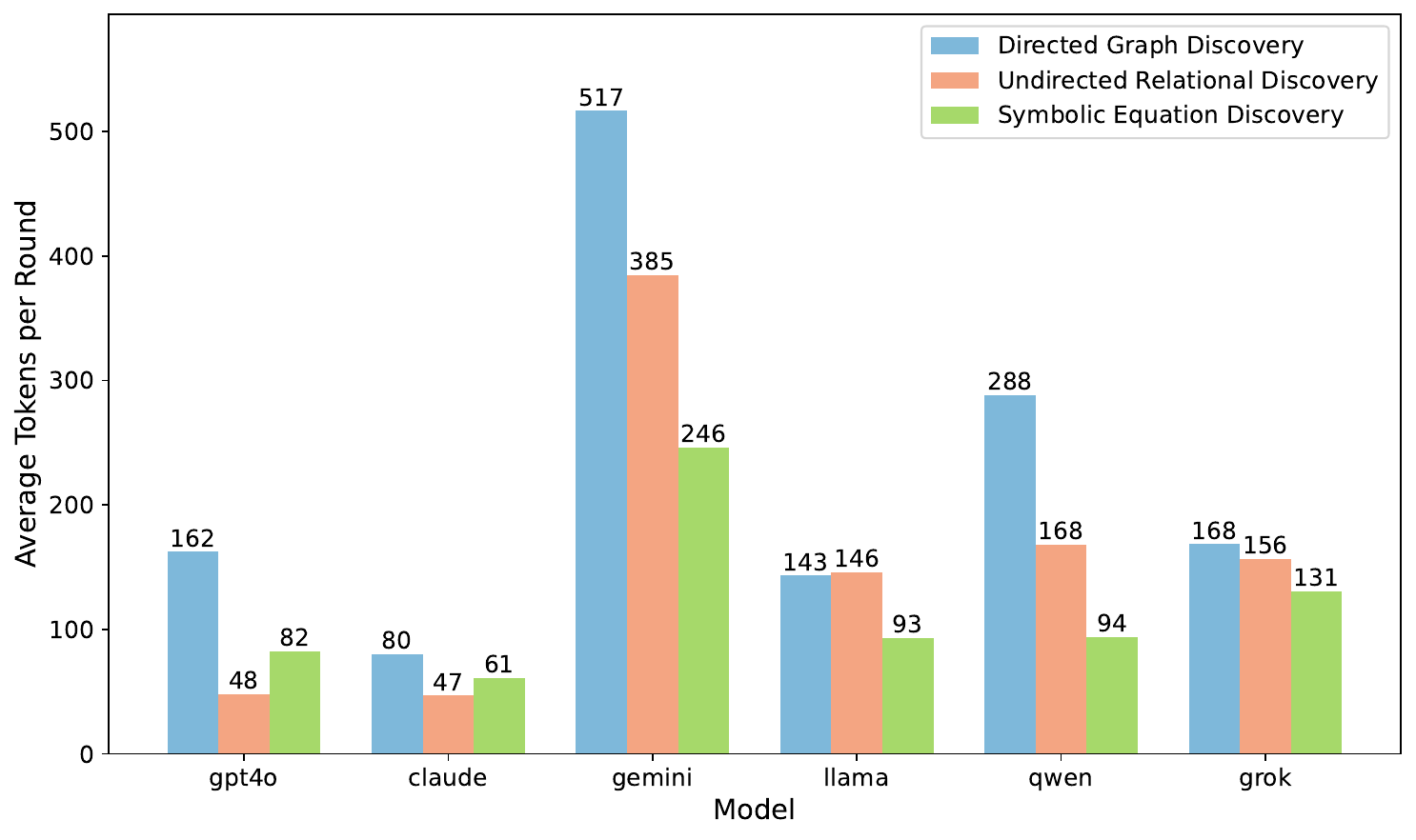}
        \caption{Chain-of-Thought Reasoning Length Comparison}
        \label{length}
    \end{subfigure}
    \vspace{-2mm}
    \caption{Chain-of-Thought reasoning analysis. (a) Keyword mentions per round across three categories: causal reasoning (green), uncertainty (orange), and evidence-based reasoning (blue). (b) Average reasoning
  length comparison across different models, measured in tokens per round.}
    \label{cot}
\end{figure*}
%%%%%%%%%%%%%%%%%%%%%%%%%%%%%

We evaluate several state-of-the-art models, including Gemini 2.5 Pro \cite{comanici2025gemini}, Grok-4~\cite{xAI2025Grok4ModelCard}, Claude-3-5 \cite{claude3.5_haiku}, GPT-4o \cite{gpt4o}, Llama-3.1 (70B) \cite{llama3.1_70b}, and Qwen2.5 (72B) \cite{qwen2.5_72b}. All experiments are conducted on the OpenRouter platform \cite{openrouter}, and the experimental budget is 365 USD. To ensure robust results and to minimize the impact of random fluctuations, we conduct 20 independent trials for directed graph discovery and 10 trials for undirected relational discovery. We will release the complete codebase upon acceptance. Below, we provide details of the models, experimental setup, and parameter configurations.

\textbf{Directed graph discovery} We evaluate three configurations, $(N,S)\in\{(3,3),(3,5),(10,5)\}$, to probe how LLM reasoning scales with graph complexity. By increasing nodes and state diversity, we test whether performance remains stable and identify thresholds where it degrades.

\textbf{Undirected relational discovery} We evaluate three network sizes—3, 5, and 10 houses to test how LLM reasoning scales with increasing relational complexity. As the number of houses and potential ties grows, we assess whether performance remains robust on larger, more intricate networks.

\textbf{Symbolic equation discovery} We evaluate five settings with varying numbers of variables (Var), three levels of formula complexity (Comp), and different numbers of distractors (Dis). By varying these settings, we test the performance of different LLMs under increasing difficulty.

\subsection{Analysis}
The results are reported in Table~\ref{Benchmark}. For directed graph discovery and undirected relational discovery, each episode is capped at \(2N\) interventions, where \(N\) denotes the number of nodes/houses. For symbolic equation discovery, the intervention budget is set to 10. If the model fails to recover the underlying ground-truth solution within the budget, we mark the episode as a failure. The success rate is the fraction of episodes solved, and the average number of iterations is computed over successful episodes only. We analyze the results from several angles. If the success rate is 0 (i.e., none of the episodes is solved), we report the average number of iterations as \(\infty\).

\noindent \textbf{Directed graph discovery}
As shown in Table~\ref{Benchmark}, GPT-4o and Qwen2.5 perform best in the small-graph regimes, achieving 100\% success for both (3 nodes, 3 states) and (3 nodes, 5 states), with similar iteration budgets (GPT-4o: 4.15 vs.\ 4.35; Qwen2.5: 4.15 vs.\ 4.45). Grok-4 and Gemini-2.5-pro are also strong (90\% success in both settings), while Claude-3.5-haiku and Llama-3.1 lag behind (success $\leq$ 35\%).

As expected, increasing the number of nodes substantially increases difficulty. In the (10 nodes, 5 states) setting, only Grok-4 and Gemini-2.5-pro retain perfect success (100\%) with around 11 iterations on average (11.0 and 11.5, respectively). In contrast, GPT-4o drops sharply to 15\% (10.67 iterations over successful episodes), while Claude-3.5-haiku, Llama-3.1, and Qwen2.5 all fail completely (0\%). These results suggest that while several LLMs handle richer state dynamics on small directed graphs, robust discovery on larger directed structures remains a major open challenge.

\noindent \textbf{Undirected relational discovery}
Table~\ref{Benchmark} shows that GPT-4o, Qwen2.5, Grok-4, and Gemini-2.5-pro all achieve 100\% success on the 3- and 5-house settings, whereas Claude-3.5-haiku and Llama-3.1 remain substantially weaker (Claude: 20\% $\rightarrow$ 0\%; Llama: 70\% $\rightarrow$ 30\%). As the network size increases to 10 houses, performance diverges sharply: Grok-4 and Gemini-2.5-pro maintain 100\% success (10.40 and 10.10 average iterations, respectively), while GPT-4o drops to 30\% and Qwen2.5 collapses to 0\%. This indicates that scaling to larger relational graphs remains challenging for several models, while the strongest models can still reliably extract and reason over the underlying structure.

In terms of efficiency, the average number of iterations increases with graph size for models that continue to succeed (e.g., Grok-4: 3.10 $\rightarrow$ 5.00 $\rightarrow$ 10.40; Gemini-2.5-pro: 3.40 $\rightarrow$ 5.30 $\rightarrow$ 10.10), consistent with the need for more probing interventions to disambiguate dense relations. Overall, current LLMs exhibit a clear scaling bottleneck on larger undirected relational graphs, suggesting that improved structural priors, better handling of long relational contexts, and more reliable multi-step reasoning could be crucial for further progress.

\noindent \textbf{Symbolic equation discovery}
This task is the most polarized one across models. Grok-4 and Gemini-2.5-pro dominate across nearly all settings: both achieve 100\% success for Var=3 with Simple/Medium/Complex formulas when no distractors are present (Dis=0), typically requiring only 1--2 iterations (e.g., Grok-4: 1.00--1.30; Gemini-2.5-pro: 1.00--2.10). When distractors are introduced (Var=3, Complex, Dis=2), Grok-4 remains at 100\% success (1.40 iterations) and Gemini-2.5-pro remains high at 90\% (1.89 iterations). The hardest setting (Var=5, Complex, Dis=2) reveals the key scaling challenge: Grok-4 retains 90\% success (3.22 iterations), while Gemini-2.5-pro drops to 60\% (4.00 iterations). In contrast, all other models stay at or below 30\% in the easiest cases and largely fail (0\%) once distractors or additional variables are introduced (often reported as $\infty$ iterations due to zero success). Overall, larger variable counts and distractors are the main drivers of difficulty, and strong symbolic reasoning appears concentrated in the top-performing models.

\noindent \textbf{Summary across three tasks}
In the three tasks, all models exhibit clear scaling effects as the problem size/complexity increases (more nodes/houses, more states, more variables, and additional distractors). Among all evaluated models, Grok-4 and Gemini-2.5-pro achieve the most consistently high success rates. However, average iterations provides a more fine-grained view beyond success rate: even when both models solve the same setting, Grok-4 typically requires fewer interventions than Gemini-2.5-pro (e.g., Symbolic equation discovery often solves in $\sim$1--3 iterations for Grok-4 versus $\sim$1--4 for Gemini-2.5-pro), indicating higher efficiency in reaching the correct solution. Overall, success rate captures whether a model can solve the task within budget, while average iterations further distinguishes how efficiently it does so.

\subsection{Reasoning Traces} 
Figure~\ref{keyword} shows reasoning-keyword frequencies in symbolic equation discovery for relationship, uncertainty, and evidence terms. Gemini-2.5-pro uses the most uncertainty and evidence terms, suggesting a more cautious, data-driven style, whereas Llama-3.1-70b-Instruct shows the lowest uncertainty usage and Claude-3.5-haiku remains consistently concise across categories. Figure~\ref{length} reports average tokens per round: Gemini-2.5-pro is the most verbose, Claude-3.5-haiku is the most concise, and GPT-4o shows task-dependent verbosity. Overall, keyword usage and response length help characterize model reasoning styles and partially align with performance trends.

\section{Evaluation on Trajectory Tracking}
The main benchmark evaluates end-to-end discovery, where success depends on intervention selection, feedback interpretation, belief maintenance, and hypothesis revision. 
To isolate structured state tracking, we introduce a trajectory-tracking diagnostic in which the model receives the full observation matrix and only identifies which variables change across consecutive steps. 
Failures in this setting therefore point to difficulty maintaining and transforming structured observations over long horizons, rather than exploration strategy alone.
\label{Trajectory}
\subsection{Task Definition}
In each round, the LLM is given an observation matrix $X \in \{0,\ldots,P\}^{M \times N}$, where $M$ is the trajectory length, $N$ is the number of nodes, and $P$ denotes the number of color states. The goal is to infer a binary change-indicator matrix $Y \in \{0,1\}^{(M-1)\times N}$, where $Y_{i,j}=1$ iff node $j$ changes color between consecutive steps $i$ and $i{+}1$, i.e., $Y_{i,j}=\mathbbm{1}[X_{i,j}\neq X_{i+1,j}]$ for $i=0,\ldots,M{-}2$ and $j=0,\ldots,N{-}1$. To reduce the impact of outliers, we repeat this procedure for $R$ rounds.

%%%%%%%%% Table 2 %%%%%%%%%%%%
\begin{table*}[t!]
\centering
\caption{Overall accuracy comparison of multiple LLMs. We report performance across various trajectory lengths, both with and without CoT prompting.}
\begin{adjustbox}{max width=0.9\textwidth}
\begin{tabular}{ccccc}
\hline
\multirow{2}{*}{\begin{tabular}[c]{@{}c@{}}Trajectory Length\\ \end{tabular}} & \multicolumn{4}{c}{Model used for evaluation (W/O CoT \textbar\ With CoT)} \\ \cline{2-5} 
& Claude-3-5-haiku & Llama-3.1-70b-instruct & GPT-4o & Qwen2.5-72b-instruct \\ \hline\hline
3  & 69\% \textbar\ 52\% & 6\% \textbar\ 8\% & \textbf{41\% \textbar\ 99\%} & \textbf{6\% \textbar\ 30\%} \\
5  & 26\% \textbar\ 29\% & 0\% \textbar\ 1\% & \textbf{13\% \textbar\ 100\%} & \textbf{1\% \textbar\ 31\%} \\
10 & 0\% \textbar\ 0\%   & 0\% \textbar\ 0\% & \textbf{0\% \textbar\ 91\%}  & \textbf{0\% \textbar\ 30\%} \\
15 & 1\% \textbar\ 0\%   & 0\% \textbar\ 0\% & \textbf{0\% \textbar\ 28\%}  & 0\% \textbar\ 5\% \\
20 & 0\% \textbar\ 0\%   & 0\% \textbar\ 0\% & 0\% \textbar\ 0\%            & 0\% \textbar\ 3\% \\
25 & 0\% \textbar\ 0\%   & 0\% \textbar\ 0\% & 0\% \textbar\ 1\%            & 0\% \textbar\ 0\% \\
30 & 0\% \textbar\ 0\%   & 0\% \textbar\ 0\% & 0\% \textbar\ 0\%            & 0\% \textbar\ 0\% \\ \hline
\end{tabular}
\end{adjustbox}
\label{Acc_vs_NoO}
\vspace{-2mm}
\end{table*}
%%%%%%%%%%%%%%%%%%%%%%%%%%%%%

%%%%%%%%% Figure %%%%%%%%%%%%
\begin{figure*}[t]
    \centering
    \begin{subfigure}{0.49\textwidth}
        \centering
        \includegraphics[width=\linewidth]{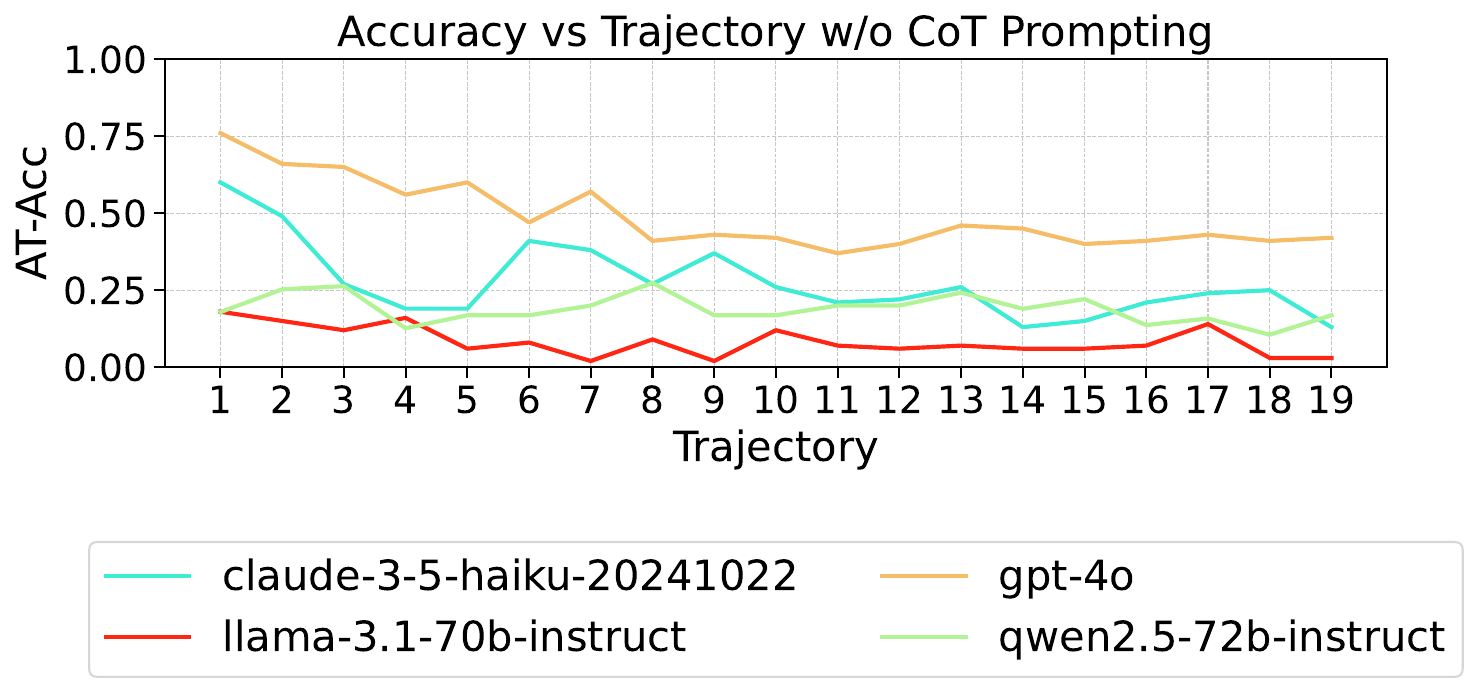}
        \caption{Without CoT Prompting}
        \label{fig:Acc_vs_Pairs_NR}
    \end{subfigure}
    \hfill
    \begin{subfigure}{0.49\textwidth}
        \centering
        \includegraphics[width=\linewidth]{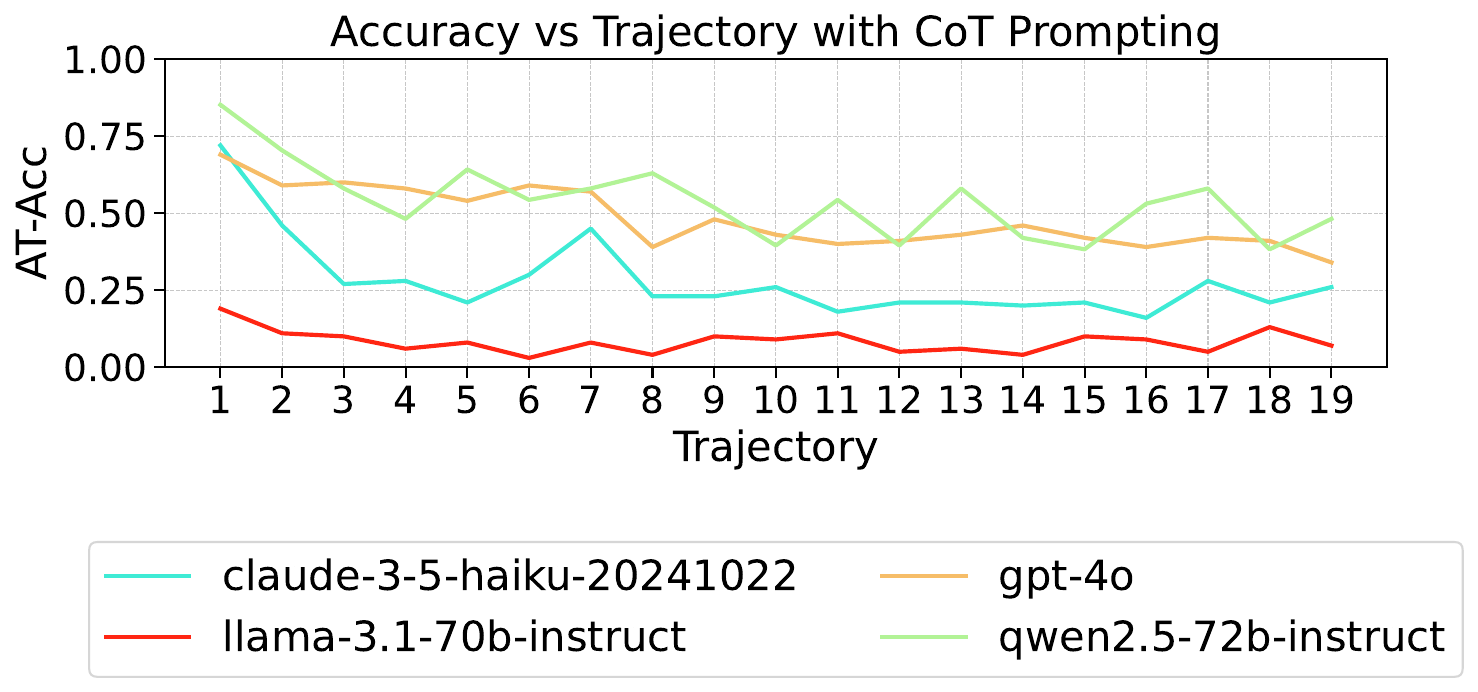}
        \caption{With CoT Prompting}
        \label{fig:Acc_vs_Pairs_R}
    \end{subfigure}
    \caption{Average trajectory accuracy across trajectory positions.}
    \label{Acc_vs_Pairs}
\vspace{-5mm}
\end{figure*}
%%%%%%%%%%%%%%%%%%%%%%%%%%%%%

\subsection{Prompting and Model Evaluation}
To assess LLMs' ability to comprehend and reason over structured information over long trajectories, we employ a structured zero-shot prompting strategy. The prompt explicitly describes the task, including matrix interpretation and expected output format. We evaluate multiple LLMs, including GPT-4o, Gemini, and Qwen, under identical conditions. Models are queried with the prompt and required to generate structured JSON outputs containing only the change-indicator matrix $Y$.

We conduct systematic evaluations across various trajectory lengths ($M \in \{3,5,10,15,20,25,30\}$) to measure the robustness of each model. To evaluate the performance of each LLM, we develop two types of metrics, denoted as Average Trajectory Accuracy (AT-Acc) and Overall Accuracy (OA-Acc). 

Overall Accuracy is computed as the fraction of exactly predicted change-indicator matrices:
\begin{equation}
    \text{OA-Acc} = \frac{1}{R} \sum_{i=1}^{R} \mathbbm{1} (Y_i = \hat{Y}_i)
\end{equation}
where $\hat{Y}$ represents the model-generated change-indicator matrix, ${Y}$ represents the ground-truth change-indicator matrix, $R$ represents the number of rounds, and $\mathbbm{1}$ is the indicator function. In our experiment, we use $R = 100$. 

Average Trajectory Accuracy measures how accurately the model predicts color changes between each consecutive pair of rows (we refer to the change from row $i$ to row $i{+}1$ as the $(i{+}1)$-th trajectory). For each round $r \in \{0,\ldots,R{-}1\}$, let $Y^{(r)}$ and $\hat{Y}^{(r)}$ denote the ground-truth and predicted change-indicator matrices, respectively. We construct $T \in \{0,1\}^{R \times (M-1)}$ where $T_{r,i}=\mathbb{I}\!\left[Y^{(r)}_{i,:}=\hat{Y}^{(r)}_{i,:}\right]$ for $i \in \{0,\ldots,M{-}2\}$, i.e., $T_{r,i}=1$ iff the entire $i$-th trajectory (all nodes) is predicted correctly in round $r$. The Average Trajectory Accuracy is then computed as:
\begin{equation}
    \begin{aligned}
        \text{AT-Acc}_j &= \frac{1}{R} \sum_{i=1}^{R} T_{i,j}, \\
        &\forall j \in \{1, \dots, M-1\}
    \end{aligned}
\end{equation}

\subsection{Chain-of-Thought Prompting}
To test whether CoT aids long-horizon tracking, we prepend the instruction \textit{``Please also include the reason for your answer''} before the prediction.

\subsection{Experiment Results} 
In this section, we present and analyze the results of our long trajectory measurement. The experimental results are presented in Table~\ref{Acc_vs_NoO} and Figure~\ref{Acc_vs_Pairs}. We analyze these results from two perspectives: 

\noindent\textbf{Overall accuracy vs.\ trajectory length.}
Table~\ref{Acc_vs_NoO} shows that accuracy drops sharply as trajectories grow longer. For example, Claude falls from 69\% at length~3 to near-zero accuracy by length~10. CoT generally helps on \emph{short} sequences (e.g., GPT-4o: 13\% $\to$ 100\% at length~5; Qwen2.5 shows similar gains), but the benefit diminishes on longer inputs, indicating limited robustness to extended evidence.

\noindent\textbf{Average trajectory accuracy across positions.}
With length fixed at 20, Figure~\ref{Acc_vs_Pairs} averages 100 trials and shows a monotonic decline in accuracy with later trajectory positions, evidencing \emph{temporal attention decay}. CoT partially mitigates this for GPT-4o and Qwen2.5, yet performance at late steps (e.g., Qwen2.5 at position~19 $<70\%$) remains weak. Future work should strengthen long-horizon reasoning (memory/attention over extended evidence) rather than relying solely on CoT.

\section{Conclusion} 
We introduced \textit{Auto-Discovery-Bench}, a controlled oracle-guided diagnostic benchmark for evaluating structured belief maintenance in closed-loop discovery. 
The benchmark includes three controlled archetypes: directed graph discovery, undirected relational discovery, and symbolic equation discovery. 
Our results reveal clear scaling limitations as problem size and complexity increase, with only the strongest models achieving consistently high success. 
Beyond success rate, average iterations provides a finer-grained measure of interaction efficiency, distinguishing models that solve the same instances but require different intervention budgets. 
Together, the benchmark and trajectory-tracking diagnostic provide reproducible, ground-truth evaluation of long-horizon structured state tracking and highlight the need for improved evidence integration and reliable multi-step inference.

\clearpage
\section{Limitations}
Auto-Discovery-Bench is not intended to simulate real scientific practice in full fidelity. 
The directed graph, undirected relational, and symbolic equation tasks are controlled abstractions with discrete states, deterministic feedback, simplified dynamics, and known ground truth. 
They omit important aspects of real discovery, including noisy measurements, partial observability, continuous variables, high-dimensional observations, non-stationary systems, and domain-specific background knowledge.

The conclusions should therefore be interpreted narrowly: the benchmark diagnoses whether models can maintain and update structured hypotheses in clean oracle-guided discovery loops. 
It does not establish that a model is capable of real-world scientific discovery, nor does failure necessarily identify a single causal mechanism. 
Although the trajectory-tracking diagnostic provides evidence that long-horizon structured state tracking is an important bottleneck, planning, representation format, prompt sensitivity, and tool use may also contribute.

Future versions can introduce controlled noise, missing observations, richer equation grammars, continuous variables, larger graph families, and standardized fixed test sets to bridge diagnostic abstraction and realistic discovery environments.

\bibliography{main}

\begin{thebibliography}{27}
\providecommand{\natexlab}[1]{#1}

\bibitem[{AI(2024)}]{llama3.1_70b}
Meta AI. 2024.
\newblock \href {https://ai.meta.com/blog/llama3-1/} {Llama 3.1: Open foundation and fine-tuned chat models}.
\newblock Model Release Notes.

\bibitem[{Altm{\"a}e et~al.(2023)Altm{\"a}e, Sola-Leyva, and Salumets}]{altmae2023artificial}
Signe Altm{\"a}e, Alberto Sola-Leyva, and Andres Salumets. 2023.
\newblock Artificial intelligence in scientific writing: a friend or a foe?
\newblock \emph{Reproductive BioMedicine Online}, 47(1):3--9.

\bibitem[{Anthropic(2024)}]{claude3.5_haiku}
Anthropic. 2024.
\newblock \href {https://www.anthropic.com/news/claude-3-5-haiku} {Claude 3.5 haiku model card}.
\newblock Accessed: 2024-10-22.

\bibitem[{Bai et~al.(2024)Bai, Bai, Chu, Cui, Dang, Deng, Fan, Ge, Han, Huang, Hui, Ji, Li, Lin, Lin, Liu, Liu, Lu, Lu, Ma, Ren, Ren, Tan, Tan, Tu, Wang, Wang, Wang, Wu, Xu, Xu, Yang, Yang, Yang, Yang, Yao, Yu, Yuan, Yuan, Zhang, Zhang, Zhang, Zhang, Zhou, Zhou, Zhou, and Zhu}]{qwen2.5_72b}
Jinze Bai, Shuai Bai, Yunfei Chu, Zeyu Cui, Kai Dang, Xiaodong Deng, Yang Fan, Wenbin Ge, Yu~Han, Fei Huang, Binyuan Hui, Luo Ji, Mei Li, Junyang Lin, Runji Lin, Dayiheng Liu, Gao Liu, Chengqiang Lu, Keming Lu, and 28 others. 2024.
\newblock \href {https://arxiv.org/abs/2405.xxxxx} {Qwen2.5: The next generation of large language models with hybrid scaling}.
\newblock \emph{arXiv preprint}.

\bibitem[{Chen et~al.(2025)Chen, Pi{\k{e}}kos, Ostaszewski, Laakom, and Schmidhuber}]{chen2025physgym}
Yimeng Chen, Piotr Pi{\k{e}}kos, Mateusz Ostaszewski, Firas Laakom, and J{\"u}rgen Schmidhuber. 2025.
\newblock \href {https://openreview.net/forum?id=w8uII2qAmd} {Physgym: Benchmarking {LLM}s in interactive physics discovery with controlled priors}.
\newblock In \emph{The Thirty-ninth Annual Conference on Neural Information Processing Systems Datasets and Benchmarks Track}.

\bibitem[{Choi et~al.(2022)Choi, Cundy, Srivastava, and Ermon}]{choi2022lmpriors}
Kristy Choi, Chris Cundy, Sanjari Srivastava, and Stefano Ermon. 2022.
\newblock Lmpriors: Pre-trained language models as task-specific priors.
\newblock \emph{arXiv preprint arXiv:2210.12530}.

\bibitem[{Cobbe et~al.(2021)Cobbe, Kosaraju, Bavarian, Chen, Jun, Kaiser, Plappert, Tworek, Hilton, Nakano et~al.}]{GSM8K}
Karl Cobbe, Vineet Kosaraju, Mohammad Bavarian, Mark Chen, Heewoo Jun, Lukasz Kaiser, Matthias Plappert, Jerry Tworek, Jacob Hilton, Reiichiro Nakano, and 1 others. 2021.
\newblock Training verifiers to solve math word problems.
\newblock \emph{arXiv preprint arXiv:2110.14168}.

\bibitem[{Comanici et~al.(2025)Comanici, Bieber, Schaekermann, Pasupat, Sachdeva, Dhillon, Blistein, Ram, Zhang, Rosen et~al.}]{comanici2025gemini}
Gheorghe Comanici, Eric Bieber, Mike Schaekermann, Ice Pasupat, Noveen Sachdeva, Inderjit Dhillon, Marcel Blistein, Ori Ram, Dan Zhang, Evan Rosen, and 1 others. 2025.
\newblock Gemini 2.5: Pushing the frontier with advanced reasoning, multimodality, long context, and next generation agentic capabilities.
\newblock \emph{arXiv preprint arXiv:2507.06261}.

\bibitem[{Dinu et~al.(2024)Dinu, Leoveanu-Condrei, Holzleitner, Zellinger, and Hochreiter}]{dinu2024symbolicai}
Marius-Constantin Dinu, Claudiu Leoveanu-Condrei, Markus Holzleitner, Werner Zellinger, and Sepp Hochreiter. 2024.
\newblock Symbolicai: A framework for logic-based approaches combining generative models and solvers.
\newblock \emph{arXiv preprint arXiv:2402.00854}.

\bibitem[{Hayes et~al.(2025)Hayes, Rao, Akin, Sofroniew, Oktay, Lin, Verkuil, Tran, Deaton, Wiggert et~al.}]{hayes2025simulating}
Thomas Hayes, Roshan Rao, Halil Akin, Nicholas~J Sofroniew, Deniz Oktay, Zeming Lin, Robert Verkuil, Vincent~Q Tran, Jonathan Deaton, Marius Wiggert, and 1 others. 2025.
\newblock Simulating 500 million years of evolution with a language model.
\newblock \emph{Science}, page eads0018.

\bibitem[{Jansen et~al.(2024)Jansen, C{\^o}t{\'e}, Khot, Bransom, Mishra, Majumder, Tafjord, and Clark}]{jansen2024discoveryworld}
Peter Jansen, Marc-Alexandre C{\^o}t{\'e}, Tushar Khot, Erin Bransom, Bhavana~Dalvi Mishra, Bodhisattwa~Prasad Majumder, Oyvind Tafjord, and Peter Clark. 2024.
\newblock \href {https://openreview.net/forum?id=cDYqckEt6d} {Discoveryworld: A virtual environment for developing and evaluating automated scientific discovery agents}.
\newblock In \emph{The Thirty-eight Conference on Neural Information Processing Systems Datasets and Benchmarks Track}.

\bibitem[{Jiralerspong et~al.(2024)Jiralerspong, Chen, More, Shah, and Bengio}]{jiralerspong2024efficient}
Thomas Jiralerspong, Xiaoyin Chen, Yash More, Vedant Shah, and Yoshua Bengio. 2024.
\newblock Efficient causal graph discovery using large language models.
\newblock \emph{arXiv preprint arXiv:2402.01207}.

\bibitem[{Jumper et~al.(2021)Jumper, Evans, Pritzel, Green, Figurnov, Ronneberger, Tunyasuvunakool, Bates, {\v{Z}}{\'\i}dek, Potapenko et~al.}]{jumper2021highly}
John Jumper, Richard Evans, Alexander Pritzel, Tim Green, Michael Figurnov, Olaf Ronneberger, Kathryn Tunyasuvunakool, Russ Bates, Augustin {\v{Z}}{\'\i}dek, Anna Potapenko, and 1 others. 2021.
\newblock Highly accurate protein structure prediction with alphafold.
\newblock \emph{nature}, 596(7873):583--589.

\bibitem[{K{\i}c{\i}man et~al.(2023)K{\i}c{\i}man, Ness, Sharma, and Tan}]{kiciman2023causal}
Emre K{\i}c{\i}man, Robert Ness, Amit Sharma, and Chenhao Tan. 2023.
\newblock Causal reasoning and large language models: Opening a new frontier for causality.
\newblock \emph{arXiv preprint arXiv:2305.00050}.

\bibitem[{Koncel-Kedziorski et~al.(2016)Koncel-Kedziorski, Roy, Amini, Kushman, and Hajishirzi}]{mawps}
Rik Koncel-Kedziorski, Subhro Roy, Aida Amini, Nate Kushman, and Hannaneh Hajishirzi. 2016.
\newblock Mawps: A math word problem repository.
\newblock In \emph{Proceedings of the 2016 conference of the north american chapter of the association for computational linguistics: human language technologies}, pages 1152--1157.

\bibitem[{Long et~al.(2023)Long, Schuster, and Pich{\'e}}]{long2023can}
Stephanie Long, Tibor Schuster, and Alexandre Pich{\'e}. 2023.
\newblock Can large language models build causal graphs?
\newblock \emph{arXiv preprint arXiv:2303.05279}.

\bibitem[{Lu et~al.(2024)Lu, Lu, Lange, Foerster, Clune, and Ha}]{lu2024ai}
Chris Lu, Cong Lu, Robert~Tjarko Lange, Jakob Foerster, Jeff Clune, and David Ha. 2024.
\newblock The ai scientist: Towards fully automated open-ended scientific discovery.
\newblock \emph{arXiv preprint arXiv:2408.06292}.

\bibitem[{Merchant et~al.(2023)Merchant, Batzner, Schoenholz, Aykol, Cheon, and Cubuk}]{merchant2023scaling}
Amil Merchant, Simon Batzner, Samuel~S Schoenholz, Muratahan Aykol, Gowoon Cheon, and Ekin~Dogus Cubuk. 2023.
\newblock Scaling deep learning for materials discovery.
\newblock \emph{Nature}, 624(7990):80--85.

\bibitem[{Miao et~al.(2021)Miao, Liang, and Su}]{ASDiv}
Shen-Yun Miao, Chao-Chun Liang, and Keh-Yih Su. 2021.
\newblock A diverse corpus for evaluating and developing english math word problem solvers.
\newblock \emph{arXiv preprint arXiv:2106.15772}.

\bibitem[{Narayan et~al.(2018)Narayan, Cohen, and Lapata}]{xsum}
Shashi Narayan, Shay~B Cohen, and Mirella Lapata. 2018.
\newblock Don't give me the details, just the summary! topic-aware convolutional neural networks for extreme summarization.
\newblock \emph{arXiv preprint arXiv:1808.08745}.

\bibitem[{OpenAI(2024)}]{gpt4o}
OpenAI. 2024.
\newblock \href {https://openai.com/index/gpt-4o/} {Gpt-4o system architecture}.
\newblock Technical Overview.

\bibitem[{OpenRouter(2025)}]{openrouter}
OpenRouter. 2025.
\newblock Openrouter: A unified interface for llms.
\newblock \url{https://openrouter.ai}.
\newblock Accessed: 2025-02-16.

\bibitem[{Patel et~al.(2021)Patel, Bhattamishra, and Goyal}]{SVAMP}
Arkil Patel, Satwik Bhattamishra, and Navin Goyal. 2021.
\newblock Are nlp models really able to solve simple math word problems?
\newblock \emph{arXiv preprint arXiv:2103.07191}.

\bibitem[{Pyzer-Knapp et~al.(2022)Pyzer-Knapp, Pitera, Staar, Takeda, Laino, Sanders, Sexton, Smith, and Curioni}]{pyzer2022accelerating}
Edward~O Pyzer-Knapp, Jed~W Pitera, Peter~WJ Staar, Seiji Takeda, Teodoro Laino, Daniel~P Sanders, James Sexton, John~R Smith, and Alessandro Curioni. 2022.
\newblock Accelerating materials discovery using artificial intelligence, high performance computing and robotics.
\newblock \emph{npj Computational Materials}, 8(1):84.

\bibitem[{Rajpurkar(2016)}]{rajpurkar2016squad}
P~Rajpurkar. 2016.
\newblock Squad: 100,000+ questions for machine comprehension of text.
\newblock \emph{arXiv preprint arXiv:1606.05250}.

\bibitem[{Wang(2018)}]{wang2018glue}
Alex Wang. 2018.
\newblock Glue: A multi-task benchmark and analysis platform for natural language understanding.
\newblock \emph{arXiv preprint arXiv:1804.07461}.

\bibitem[{xAI(2025)}]{xAI2025Grok4ModelCard}
xAI. 2025.
\newblock \href {https://data.x.ai/2025-08-20-grok-4-model-card.pdf} {Grok 4 model card}.
\newblock Accessed: 2025-08-25.

\end{thebibliography}

% \appendix

% \section{Example Appendix}
% \label{sec:appendix}
\clearpage
\appendix
\label{Appendix}
% \includepdf[
%   pages=-,
%   scale=0.8,
%   pagecommand={\section{Appendix}}
% ]{Fig/Prompt.pdf}

\includepdf[
  pages=1,
  scale=0.8,
  pagecommand={\twocolumn[{\section{Appendix}}]}
]{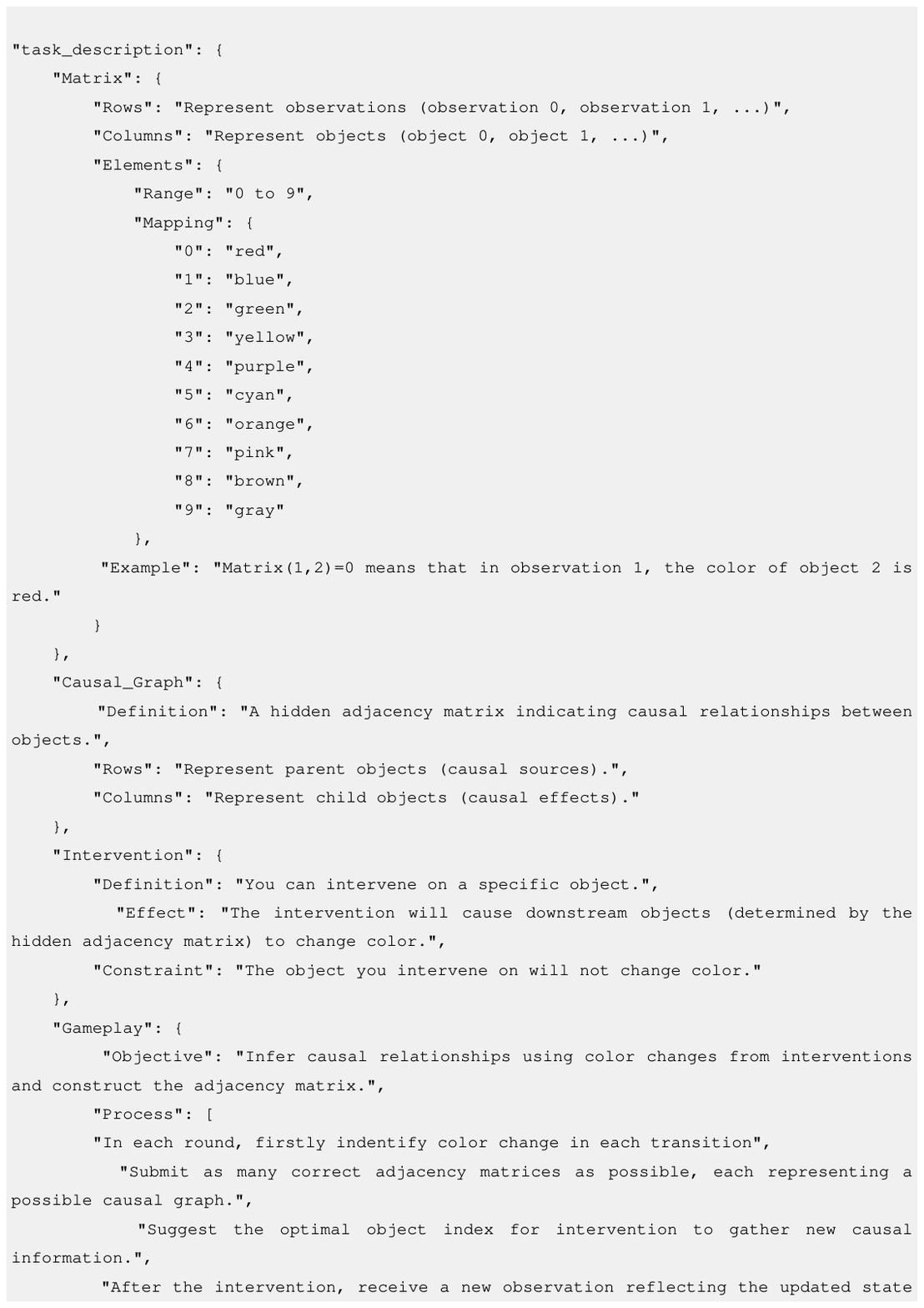}

\includepdf[
  pages=2-,
  scale=0.8,
  pagecommand={}
]{Fig/Prompt.pdf}

\end{document}